\documentclass[12pt]{spieman}
\usepackage{amsmath,amsfonts,amssymb}
\usepackage{graphicx}
\usepackage{setspace}
\usepackage{tocloft}

\title{Physics-informed network paradigm with data generation and background noise removal for diverse distributed acoustic sensing applications}

\author[1]{Yangyang Wan}
\author[1]{Haotian Wang}
\author[1]{Xuhui Yu}
\author[1]{Jiageng Chen}
\author[1]{Xinyu Fan}
\author[1,*]{Zuyuan He}

\affil[1]{State Key Laboratory of Advanced Optical Communication Systems and Networks, Department of Electronic Engineering, Shanghai Jiao Tong University, Shanghai 200240, China.}

\cftpagenumbersoff{figure}
\cftpagenumbersoff{table} 
\begin{document} 
\maketitle

\begin{abstract}
Distributed acoustic sensing (DAS) has attracted considerable attention across various fields and artificial intelligence (AI) technology plays an important role in DAS applications to realize event recognition and denoising. Existing AI models require real-world data (RWD), whether labeled or not, for training, which is contradictory to the fact of limited available event data in real-world scenarios. Here, a physics-informed DAS neural network paradigm is proposed, which does not need real-world events data for training. By physically modeling target events and the constraints of real world and DAS system, physical functions are derived to train a generative network for generation of DAS events data. DAS debackground net is trained by using the generated DAS events data to eliminate background noise in DAS data. The effectiveness of the proposed paradigm is verified in event identification application based on a public dataset of DAS spatiotemporal data and in belt conveyor fault monitoring application based on DAS time-frequency data, and achieved comparable or better performance than data-driven networks trained with RWD. Owing to the introduction of physical information and capability of background noise removal, the paradigm demonstrates generalization in same application on different sites. A fault diagnosis accuracy of 91.8\% is achieved in belt conveyor field with networks which transferred from simulation test site without any fault events data of test site and field for training. The proposed paradigm is a prospective solution to address significant obstacles of data acquisition and intense noise in practical DAS applications and explore more potential fields for DAS.
\end{abstract}

\keywords{distributed acoustic sensing, machine learning, optical fiber sensing, event recognition}

{\noindent \footnotesize\textbf{*}Zuyuan He,  \linkable{zuyuanhe@sjtu.edu.cn} }

\begin{spacing}{1}   

\section{Introduction}
Distributed acoustic sensing (DAS) offers long measurement range, high spatial resolution, high sensitivity and rapid measurements in real-time acoustic monitoring, leading to its extensive applications\cite{he2021optical,liu2022advances}.
DAS interrogator unit launches pulsed light into the optical fiber and receives the returning Rayleigh backscattering lightwave.
By demodulating the phase changes of the Rayleigh backscattering lightwave, longitudinal strain or strain rate over time caused by acoustic waves at any position along the fiber can be acquired\cite{shang2022research,lu2010distributed}.
The high performance of DAS makes it a prospective solution for applications such as pipeline monitoring, geophysical exploration, structural health monitoring, seismic surveillance, and perimeter security\cite{gorshkov2022scientific,zhang2024current,chen2023advanced,huang2020event}.
In contrast to the excellent results achieved in laboratory simulation settings\cite{li2023spatial,shi2020multi,xu2018pattern}, DAS may has a high rate of false positives or false negatives, and even fail to detect target event information in field applications.
The common main challenges across various applications of DAS lie in obtaining sufficient useful event data for signal analysis and extracting target event signals from severe environmental noise\cite{zhang2024current}.
In recent years, artificial intelligence (AI) technology has developed rapidly and has been demonstrated to be applicable for DAS signal denoising, phase unwrapping, and event identification\cite{venketeswaran2022recent,li2022ameliorated,Wu:24,wu2019dynamic,lyu2020distributed}. 
AI, with its powerful signal processing and analysis capabilities, offers potential solutions to the challenges faced in DAS practical applications\cite{tejedor2017machine,rahman2023deep,tejedor2019contextual}.


The first crucial challenge is the limited availability of events data\cite{shi2022event}.
Target events typically occur at only a few spatial locations with relative short durations.
Consequently, the vast majority of the massive spatiotemporal data provided by DAS are background data, with a limited number of target events data. 
Furthermore, labeling target events in practice is not only time-consuming and labor-intensive but also challenging to ensure label accuracy due to the enviromental noise and manual operation.
Although many supervised neural networks have demonstrated excellent performance in DAS applications\cite{huang2024multiple,dong2020random,li2024deep,yang2023denoising}, they typically require a large amount of high-quality labeled data for training, and the test data often shares the same experimental setup and environment as the training data.
Semi-supervised learning and self-supervised learning neural networks have been explored to address the problem of labeled data requirement\cite{wu2022improved,shi2021easy}.
Semi-supervised learning based neural networks for DAS data processing are employed to address the phase-picking task of DAS data for earthquake monitoring\cite{zhu2023seismic}, conduct damage detection in pipes\cite{sen2019data,yang2021long}, and perform track detection on high-speed railway\cite{wang2022semi}.
A self-supervised learning fault classification algorithm with requirement of a small number of labeled samples is proposed for fault detection in belt conveyors\cite{ZHENG2024111697}.
By leveraging large amount of unlabeled data, the classification network learns the intrinsic features associated with faults.
Unsupervised deep neural network is utilized in geoscience to extract features from DAS spatiotemporal data for the purpose of event category discrimination\cite{CHIEN2023105223}.
An adaptive decentralized artificial intelligence is proposed to improve generalization performance of DAS algorithm by fine-tuning pre-trained model with the unlabeled data in each site\cite{Shixiong}.
Although above metohds can effectively reduce or even eliminate the need for labeled data, they still require a sufficient amount of data that includes the target events for training.
In applications such as structural health monitoring and seismic detection\cite{fernandez2022seismic}, long-term data collection in the real world is still a difficult task due to the infrequent occurrence of target events.
To avoid reliance on field data for training, a generative adversarial net (GAN) based DAS data generation network is proposed to provide simulated data to serve as training dataset\cite{8725535}.
Real-world data (RWD) is still required for training GAN, and the accuracy of a three-class classification network trained only with simulation dataset is 64 \% when tested on experimental data.
Using synthetic data as labeld data is another approach to network training.
By employing finite difference modelling to simulate various microseismic events, simulated synthetic data acts as the training dataset for neural network\cite{liu2022convolutional}.
After training, an accuracy of 82 \% was achieved in binary classification on field data tests. 
It is worth noting that the network trained with seismic waveforms measured by traditional seismometers can be applied to DAS data for effective seismic waves detection\cite{9664395}, which means that both DAS and seismometer signals imply the similar physical process.

The other key challenge is the environmental noise interference to the target signals in DAS applications\cite{zhang2023modified}.
Due to the high sensitivity characteristic of DAS, acoustic signals generated by non-target events in the actual environment are received, which constitute the environmental background noise with complex components and diverse attributes.
Since conventional denoising algorithms struggle to address such complex noise, various machine learning based DAS denoising algorithms have been explored\cite{lapins2024n2n,zhong2023multiscale}.
The acquisition of training data, especially noise-free ground truth, for denoising networks is also a challenging or even impossible task.
Benefiting from the high spatial density of DAS data, self-supervised learning methods can remove spatially incoherent noise with unknown characteristics in DAS data without the need for noise-free ground truth\cite{9655039,saad2024signal,yuan2023spatial}.
Furthermore, an improved blind spot network has been proposed to remove noise with certain spatial correlation\cite{10226343}.
Although self-supervised methods reduce the dependence on noise-free ground truth, noise typically needs to meet certain conditions, such as zero mean and a certain degree of independence.
Moreover, self-supervised learning strategy may mistakes background noise signals generated by environmental interference events for fault event in applications such as perimeter security and structural health monitoring.
The key to background noise removal still lies in obtaining noise-free data of the target events.
The utilization of synthetic data as noise-free data is a approach for the training of denoising neural networks.
In traffic monitoring, an artificial algorithm is engineered to produce synthetic data of traffic trajectories, which are served as labels for the training of denoising network\cite{xie2024intelligent}.
In seismic wave monitoring applications, noise-free seismic data are generated by solving the elastic wave equation in many works, which are used as clean data for training denoising networks\cite{9684892,9579254,s23208619}.
It is worth noting that the distinctive features of differential phase signals from DAS outputs are not considered in current synthetic data schemes of seismic monitoring field.
From the above, most existing research is reliant on real-world data, and a gap is the insufficient consideration of the physical characteristics of DAS data in various applications.
Despite synthetic data schemes have been demonstrated for noise reduction and event identification in seismic wave monitoring due to mature theoretical models, most DAS applications in complex scenarios lack the method of establishing the physical model of DAS data, which hinders the broadening and employment of DAS in various applications.

%

In this work, we present a physics-informed neural network paradigm capable of generating DAS data and removing background noise for deployment in diverse DAS applications without needing real-world event data.
The proposed paradigm firstly utilizes physical functions to train the physics-informed generative network (PIGN) to generate a large amount of DAS event data, which solves the challenge of obtaining DAS events data.
The physical functions consists of the physical model and expert experience corresponding to the event, as well as the constraint functions of the real world and DAS system.
By combining the generated event data with easily obtainable real-world normal background data, a DAS debackground net for elimination of intense background noise is trained.
After training the classification network with debackground data, it can be applied in the field for event recognition.
The proposed paradigm is validated on different DAS applications.
In a event recognition task, DAS data of shake and walk events are generated by PIGN and compared with RWD in the public dataset.
The classification performance of networks trained with generated data are comparable to networks trained with RWD.
The belt conveyor fault monitoring application is selected due to the presence of significant background noise.
At the simulation test site, the debackground data from debackground net has better classification performance than initial data in various classification networks due to its effective removal of intense background noise similar to event signals.
Attributed to the introduction of physical knowledge, the network paradigm shows generalization capability and can be quickly deployed in different sites of similar DAS applications.
After transferring the network paradigm from the simulation test site to the field, a belt conveyor fault diagnosis accuracy of 91.8\% in field data without any real-world event data for training is achieved, which is higher than the results of 82.2\% and 86.6\% for the conventional data-driven neural network and artificially design algorithm based on sufficient real-world fault signals.

\section{Results}

\begin{figure*}[tb]
\centering
\includegraphics[width=16cm]{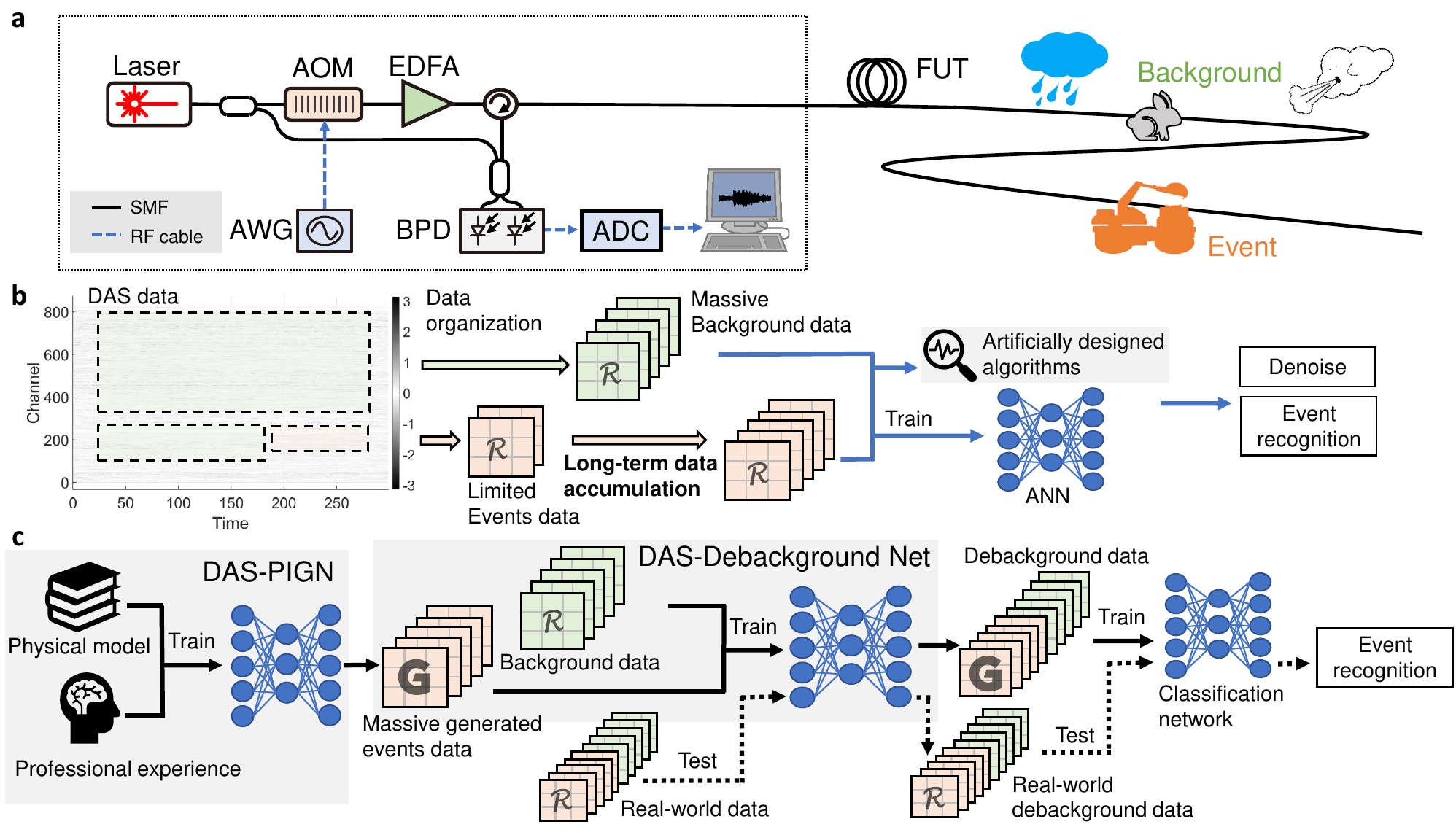}
\caption{\textbf{Schematic of the proposed physics-informed DAS neural network paradigm.} \textbf{a} Schematic diagram of DAS system and application. SMF, single-mode fiber; RF, radio frequency; AWG, arbitrary waveform generator; AOM, acousto-optic modulator; EDFA, erbium-doped optical fiber amplifer; BPD, balanced photodetector; ADC, analog-to-digital converter; FUT, fiber under test. DAS system simultaneously detects both background and event signals in practical application. \textbf{b} Traditional design process of DAS signal processing algorithms. ANN, artificial neural network. A long-term data accumulation process is required to obtain a sufficient amount of event data, whether labeled or not. \textbf{c} Physics-informed DAS neural network paradigm. PIGN, physics-informed generative network. The training of PIGN, debackground net, and classification network in the network paradigm do not require any real-world event data. PIGN is used to generate data, debackground net is used to remove background noise, and classification network is used for event recognition.}
\label{fig1}
\end{figure*}
\subsection{Physics-informed DAS neural network paradigm}
The experimental setup and application deployment of DAS system are shown in Fig.1a.
In DAS, the laser is modulated into pulsed light and injected into the fiber under test (FUT).
The location of any point along FUT can be determined according to round-trip time of pulsed light in FUT.
Rayleigh backscattering light field generated on FUT interferes with the local light field to generate intereference signal.
Since the phase change of the backscattering light field is linear with the strain rate on the optical fiber, the quantitative measurement of acoustic wave can be realized by acquiring phase information from the interference signal.
When DAS is applied to the field application, the acoustic waves of target events and background interference events are received simultaneously, which makes it difficult to extract and analyze the target signal.
For most DAS applications, the occurrence of the target events is rare, such as seismic wave detection, structural health monitoring and pipeline monitoring.
Therefore, the vast majority of DAS data belongs to background data containing various environmental noises due to the characteristic of high spatiotemporal density in DAS data.
It leads to the requirement of a long-term time accumulation process to obtain considerable target events dataset for the design of artificially designed algorithms or the training of conventional data-driven neural networks for DAS data processing, as shown in Fig.1b.
Data accumulation process may encounters difficulties when the application environment is complex and the background noise is intense.

In order to overcome the challenges of data accumulation process and significant background noise in DAS applications, physics-informed neural network paradigm including DAS data generation and denoising is proposed as shown in Fig.1c.
DAS physics-informed generative network (PIGN) is firstly proposed to produce massive generated events data.
PIGN only needs to be constrained by the physical model of the target event in the training process, without requirement of any RWD.
For events that are difficult to be physically modeled, professional experience can be converted into empirical functions as the training constraints for PIGN.
The generated data can be regarded as noise-free target event signals.
With these generated noise-free ground truth, a DAS debackground net is proposed to eliminate complex environmental noise.
The training dataset of debackground net is composed of generated events data and easily obtainable real-world background data.
Since the background environment is unique in different DAS application scenarios, it is difficult and unnecessary to physically model the background data.
In DAS practical application, DAS signals at most spatial locations belong to the background data, which is easy to obtain.
Given the characteristics of DAS background data, the real-world background data should be directly collected for data analysis.
The common core purpose in most DAS applications is to realize event recognition.
Hence, a neural network for event classification is trained using generated events data and real-world background data. 
The proposed DAS data processing network paradigm does not require real-world events data throughout the entire training process.
Once debackground net and classification network have been trained, they can be directly applied to RWD for environmental noise removal and event recognition task in the field.

\subsection{Physics-informed generative network}
The success of the proposed network paradigm hinges on the minimal difference between the generated data produced by PIGN and RWD.
The working principle of PIGN is shown in Fig.2.
Signals generated by different types of events often possess distinct spatiotemporal features.
Since DAS data is two-dimensional, comprising temporality and spatiality, it is capable of recording features of various events.
The spatial and temporal features of an event can be delineated by viewing DAS data as an image and projecting curves along the respective axes, as shown in Fig.2a.
Typically, the temporal and spatial feature functions of a class of event can be derived from theoretical physical models based on research in related fields.
For applications where the theory is underdeveloped or difficult to model due to the sophisticated circumstances, the temporal and spatial feature functions can be derived from expert knowledge.
With feature functions, massive feature curves can be obtained to depict multiple cases of the event.
In PIGN, temporal and spatial feature curves of the event are adopted as the objective curves for network training as shown in Fig.2b.
PIGN is based on U-Net structure with random data as input and generated event data as output. 
\begin{figure*}[tb]
\centering
\includegraphics[width=16cm]{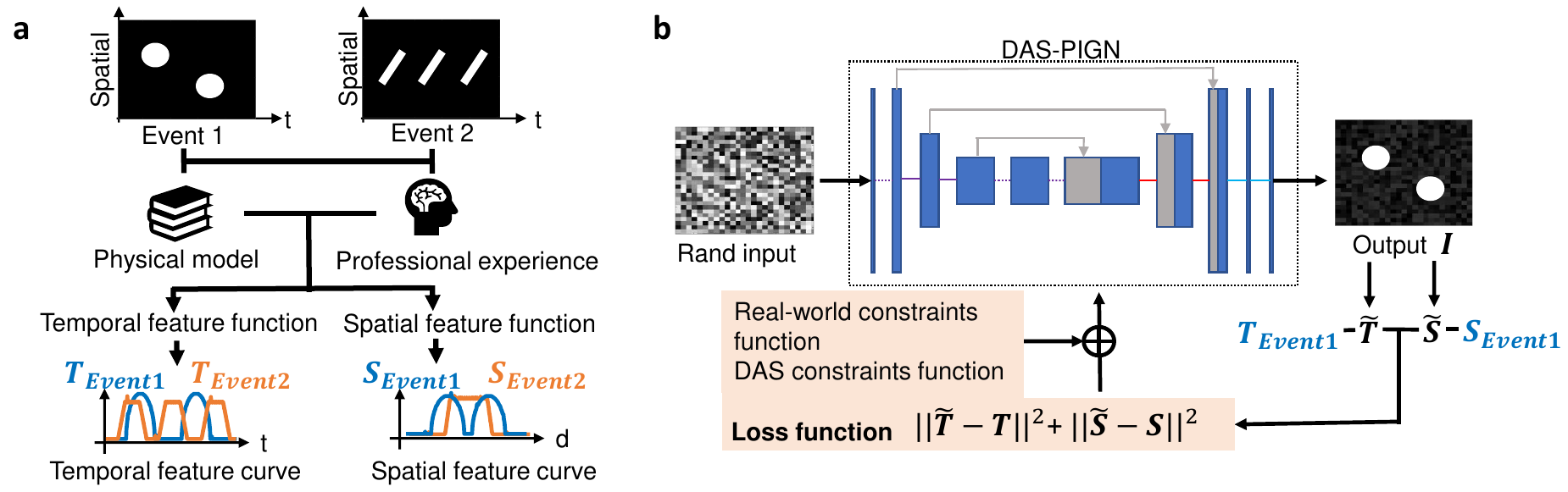}
\caption{\textbf{Principle of PIGN.} \textbf{a} Schematic of feature functions acquisition for DAS event data. After projecting DAS event data onto different dimensions, and combining physical models with professional experience, the corresponding feature functions of the events on the respective dimensions can be obtained. The feature curves of an individual event can be obtained through the feature functions. \textbf{b} Network structure and training process of PIGN. The network structure is based on the U-Net structure. During training, the loss function is established by assessing the gap between the feature curves of the output image and the target feature curves, while also accounting for constraints from DAS and real world.}
\label{fig2}
\end{figure*}
The output images are projected onto the temporal and spatial dimensions to obtain corresponding feature curves, and the squared norm of the difference between these curves and the objective curves is used as the loss function.
Although high-quality image reconstruction often requires curve projections from multiple directions\cite{smith1985image}, projection curves in the dimensions of time and space have provided sufficient information for PIGN to learn the spatiotemporal features of the event.
To narrow the gap between generated data and RWD, constraints functions of real-world and DAS system have been added to the loss function.
Since real-world signals are predominantly continuous in time domain, the mean square value of output image difference data is computed and incorporated as a real-world physical constraint function into the loss function, which can be expressed as:
\begin{equation}
\frac{1}{T}\int_0^T {{{(\frac{{I(t,s)}}{{\partial t}})}^2}} dt
\label{eq1}
\end{equation}
where $t$ is time domain, $s$ is spatital domain, $T$ is total time and $I(t,s)$ is the output image.
With this continuous signal constraint function, unreal spurious spikes in the generated data can be eliminated.
In many scenarios, correlations in DAS data can be observed across various channels within dimensions like spatial, temporal, and frequency domains.
Taking the spatial dimension as an example, spatial correlation constraint function added to the loss function can be expressed as:
\begin{equation}
{(\sum\limits_i {\sum\limits_j {{C_{{s_i}{s_j}}}} }  - {C_T})^2}
\label{eq1}
\end{equation}
where ${{C_{{s_i}{s_j}}}}$ is correlation coefficient between channel $i$ and $j$, $C_T$ is the target value of the total correlation coefficient.
Continuous signal and correlation constraint function in other dimensions can be constructed in a similar way according to the event characteristics.
Moreover, since phase wrapping is a common phenomenon and important characteristic in DAS data, a phase wrapping algorithm should be applied to the objective feature functions.
Taking the temporal feature function $T$ as an example, phase wrapping constraint can be expressed as:
\begin{equation}
{T_{wrapped}} = T - 2\pi \left[ {\frac{{T + \pi }}{{2\pi }}} \right]
\label{eq1}
\end{equation}
where $\left[  \cdot  \right]$ is floor function.
Details of constraints function of real-world and DAS system in sections 1.5 and 1.6 of supplementary document.
By using the objective feature functions as the learning target and integrating constraints of real-world and DAS, PIGN can synthesize massive DAS data for target events.
There are two training methods for PIGN: untrained and trained.
Details in Methods.
Untrained mode is suitable for the generation of small batch data with specific features, while trained mode has advantage in the generation of massive data.

\begin{figure*}[tb]
\centering
\includegraphics[width=16.5cm]{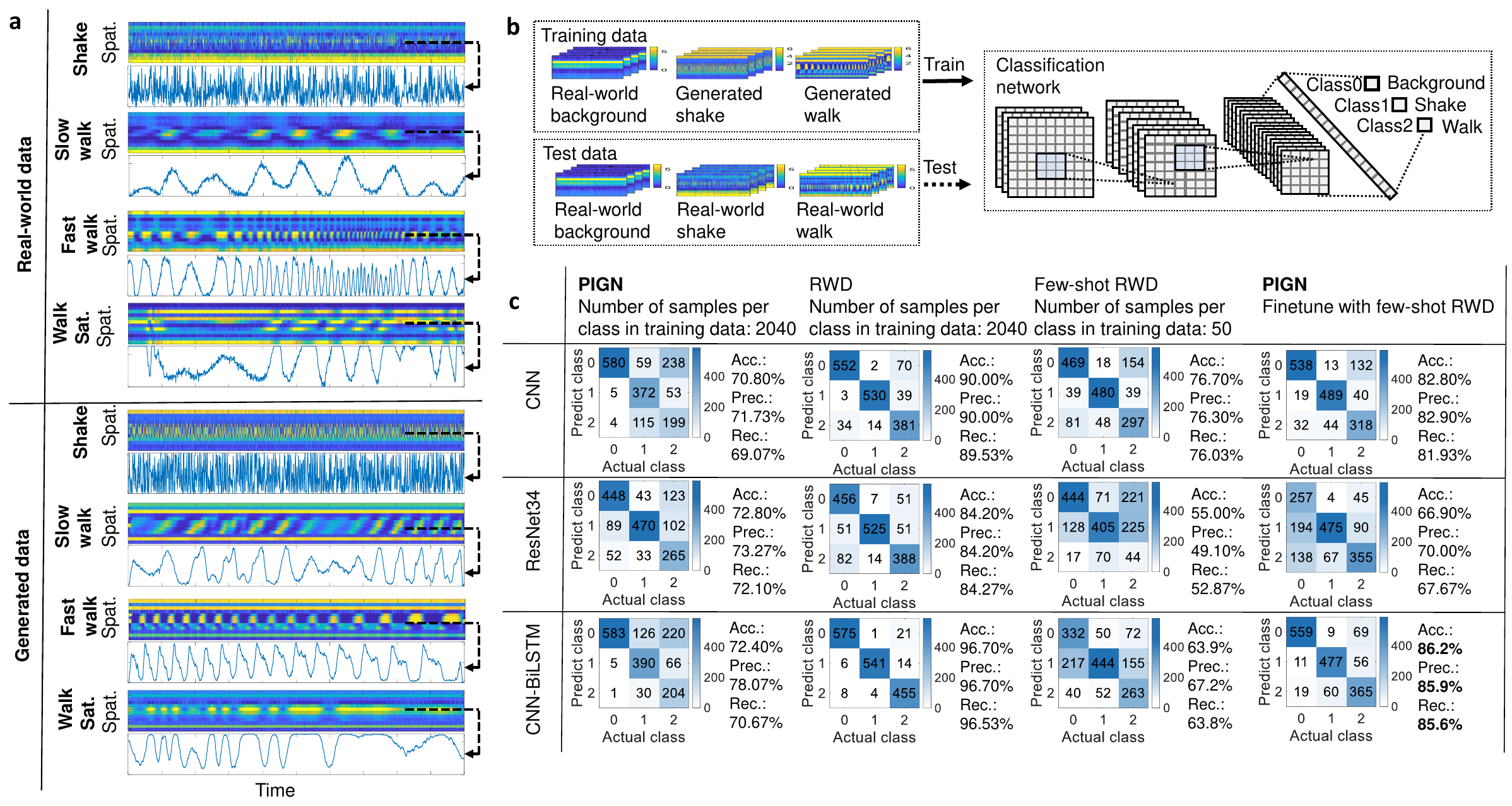}
\caption{\textbf{Event recognition performance on public dataset.} \textbf{a} Examples of shake and walk DAS data from public dataset and PIGN-generated dataset. Spat., spatial; Sat., saturation. PIGN can generate data that possesses the characteristics of signal saturation and phase wrapping observed in practical applications. \textbf{b} Flowchart of the training and testing process for the classification network. The training dataset comprises only easily accessible real-world background data and event data generated by PIGN, with no real-world event data included. \textbf{c} Event recognition results in different networks with same real-world test data from public dataset. RWD, real-world data; Acc., accuracy; Prec., precision; Rec., recall. class 0: background; class 1: shake; class 2: walk. PIGN means training with generated event data, whereas RWD means training with real-world event data.}
\label{fig3}
\end{figure*}

\subsection{Event recognition with PIGN}
The effectiveness of PIGN is validated on a public DAS dataset through event recognition task\cite{li2024deep}.
The public DAS dataset includes training and testing datasets for six distinct event classes, with each class having roughly 2000 training and 500 testing samples.
The background, shake, and walk data have been selected to validate the PIGN.
Shaking can be described by a forced vibration model, while walking conforms to a pedestrian motion model.
See in Methods.
Based on these physical models, combined with the model of DAS signal generation mechanisms, both temporal and spatial feature functions can be derived for these two classes of events.
With the objective feature functions, PIGN can generate corresponding DAS event data, which is compared with RWD as shown in Fig.3a.
The generated data and RWD present similar structural patterns.
Owing to the integration of DAS system and real-world constraint functions, the generated data can simulate signal saturation and phase wrapping.
As shown in Fig.3b, generated event data and real-world background data are used as training data to train a classification network.
After training, RWD is used for testing.
Multiple classification networks were trained with varying amounts of RWD under the same network structure and parameters for comparison with the classification network trained with generated data, as shown in Fig.3c.
Three representative basic classification networks, convolutional neural network (CNN)\cite{wu2017introduction}, residual network (ResNet)\cite{he2016deep}, and convolutional neural network - bidirectional long short-term memory (CNN-BiLSTM)\cite{lu2021cnn}, were trained to demonstrate that the validity of the generated data is unrelated to network structure.
The classification accuracies of the three types of networks trained with PIGN output data are above 70\%, surpassing the 55\% and 63.9\% accuracies of ResNet and CNN-BiLSTM networks trained with 50 samples of RWD per class.
Networks initially trained with data from PIGN were further finetuned using a small RWD dataset with 50 samples per class.
This process led to a 6.1\%, 11.9\%, and 22.3\% increase in accuracy for networks CNN, ResNet, and CNN-BiLSTM, respectively, over networks trained on same small RWD dataset. 
The finetuned CNN-BiLSTM reached an accuracy of 86.2\%, exceeding the performance of ResNet when trained on a large dataset of RWD.
The results demonstrate that the generated data produced by PIGN can effectively simulate RWD, and the classification network trained with generated data has comparable or superior performance to the networks trained with a limited amount of RWD.


\subsection{Debackground net}
\begin{figure*}[tb]
\centering
\includegraphics[width=17cm]{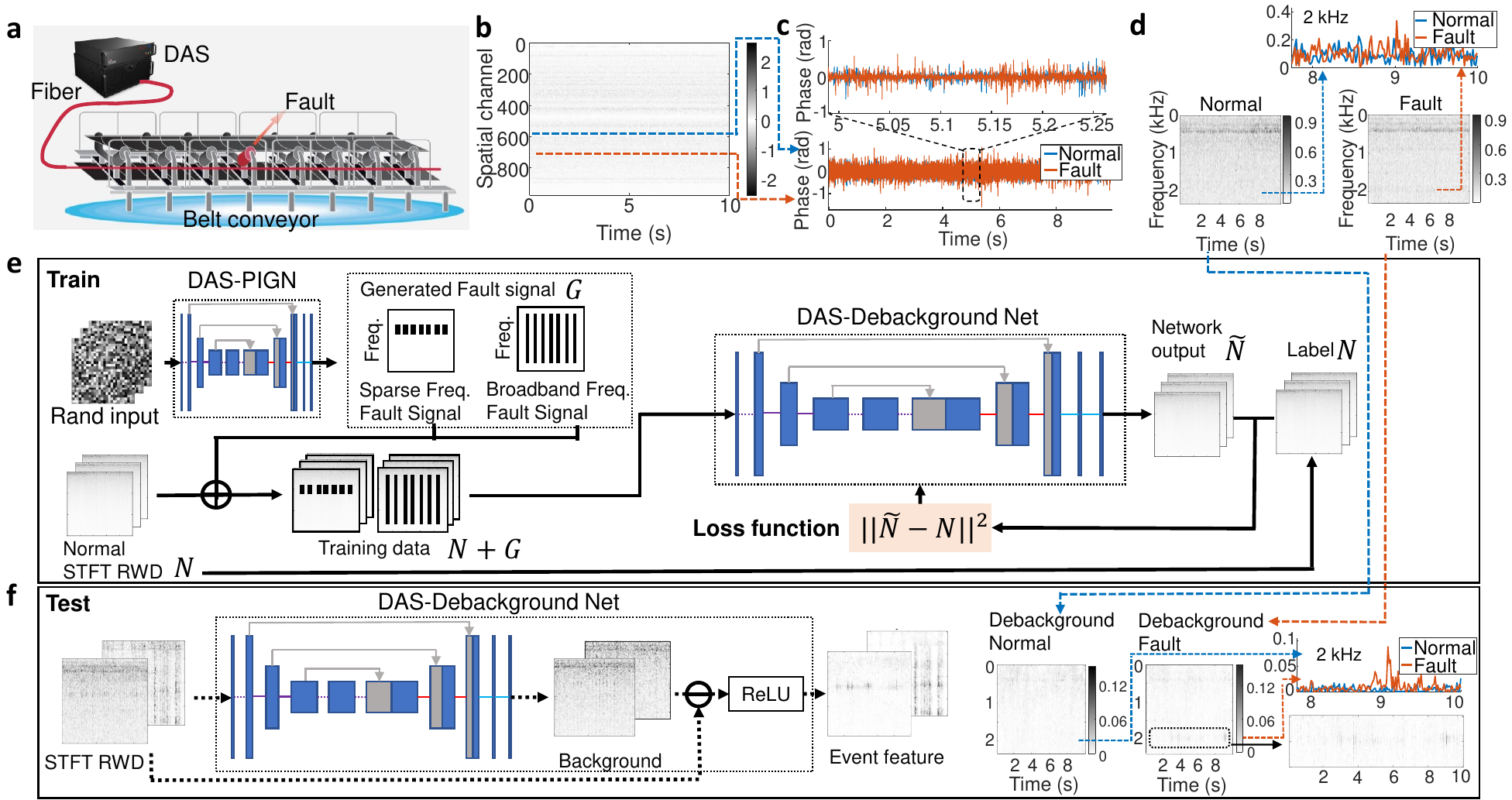}
\caption{\textbf{Schematic of the DAS debackground net.} \textbf{a} Schematic illustration of belt conveyor fault monitoring. \textbf{b} Example DAS data in belt conveyor monitoring. \textbf{c} Temporal DAS signals of normal and fault state in belt conveyor. \textbf{d} DAS time-frequency diagrams for fault and normal state. The zoomed-in figure depicts the the temporal variation of signals at 2 kHz frequency for both fault and normal states. Due to the presence of intense background noise signals, it is difficult to distinguish normal and fault signal in both time and time-frequency domain. \textbf{e} The training process of DAS debackground net. STFT, short-time fourier transform. DAS debackground net is based on the U-Net structure. \textbf{f} The test process of DAS debackground net. ReLU, rectified linear unit. After debackground, the fault signal can be distinctly observed at the 2 kHz frequency. }
\label{fig4}
\end{figure*}
\begin{figure*}[tb]
\centering
\includegraphics[width=17cm]{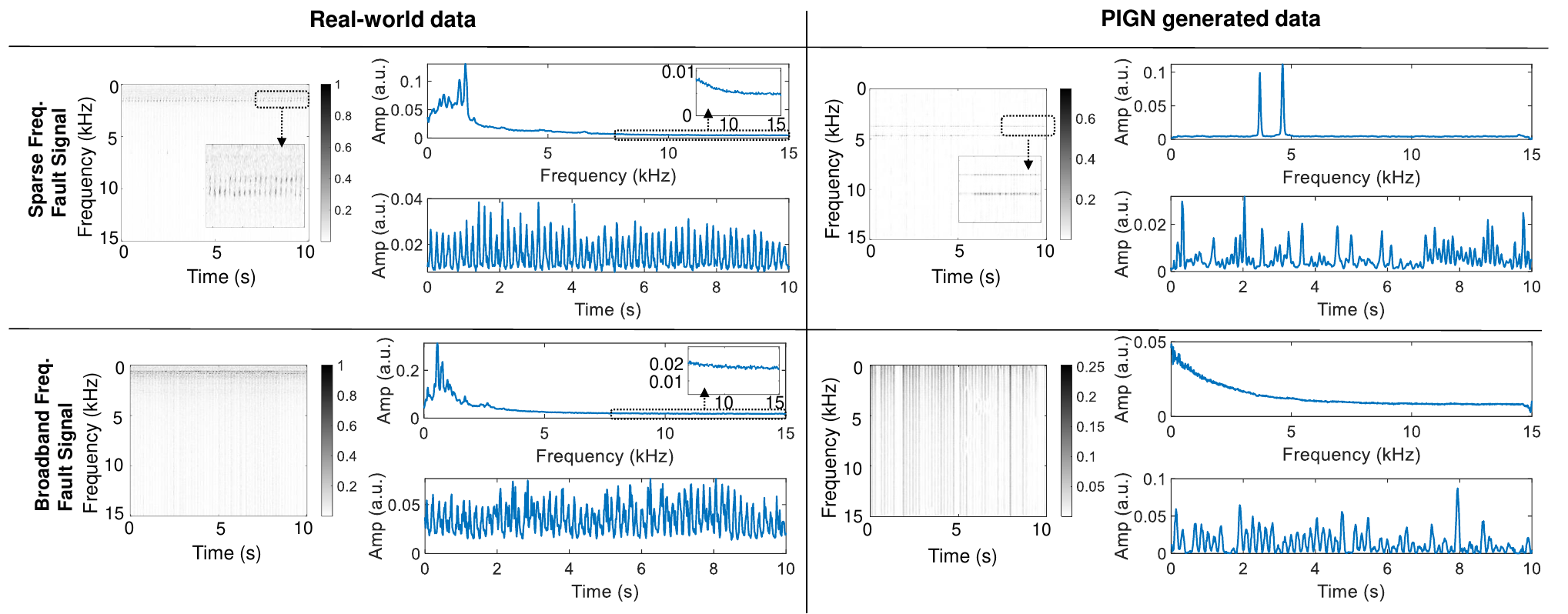}
\caption{\textbf{Examples of fault signals in belt conveyor of real-world data and PIGN-generated data.} Real-world data with intense fault signals are presented for clear demonstration, and show periodic impulsive features with varying intensities in the time domain. PIGN-generated data possesses characteristics similar to real-world data.}
\label{fig5}
\end{figure*}
Generated data from PIGN can also be utilized to train a DAS signal denoising network, addressing the challenge of severe interference of environmental background noise in DAS applications.
Due to obtaining DAS data in a laboratory environment, the background noise in the public dataset is not intense enough.
Distributed monitoring of belt conveyors is a typical DAS application, as shown in Fig.4a, with significant background noise.
Fig.4b shows a common DAS data in the temporal and spatial dimensions for belt conveyor monitoring.
Mechanical vibrations accompany the normal running of belt conveyor, causing significant background noise in DAS data. 
Due to the characteristics of intense background noise and intermittent fault signals in belt conveyor monitoring application, the time-frequency DAS data is acquired by short-time Fourier transform (STFT) of the time-domain DAS data at each spatial channel to analyze the time-varying frequency information of the signal.
For less severe faults, the relatively weak fault signals are often overwhelmed by the background noise, making it difficult to distinguish between fault and normal signals in both the time and time-frequency domains, as shown in Fig.4c and Fig.4d.
Moreover, the vibration from normal operation share similar mechanisms with the vibration induced by some types of faults, resulting in background noise in the normal signal has similar features with the fault data.
Therefore, it is necessary to performing denoising to extract the target event signal in DAS applications.

\begin{figure*}[tb]
\centering
\includegraphics[width=17cm]{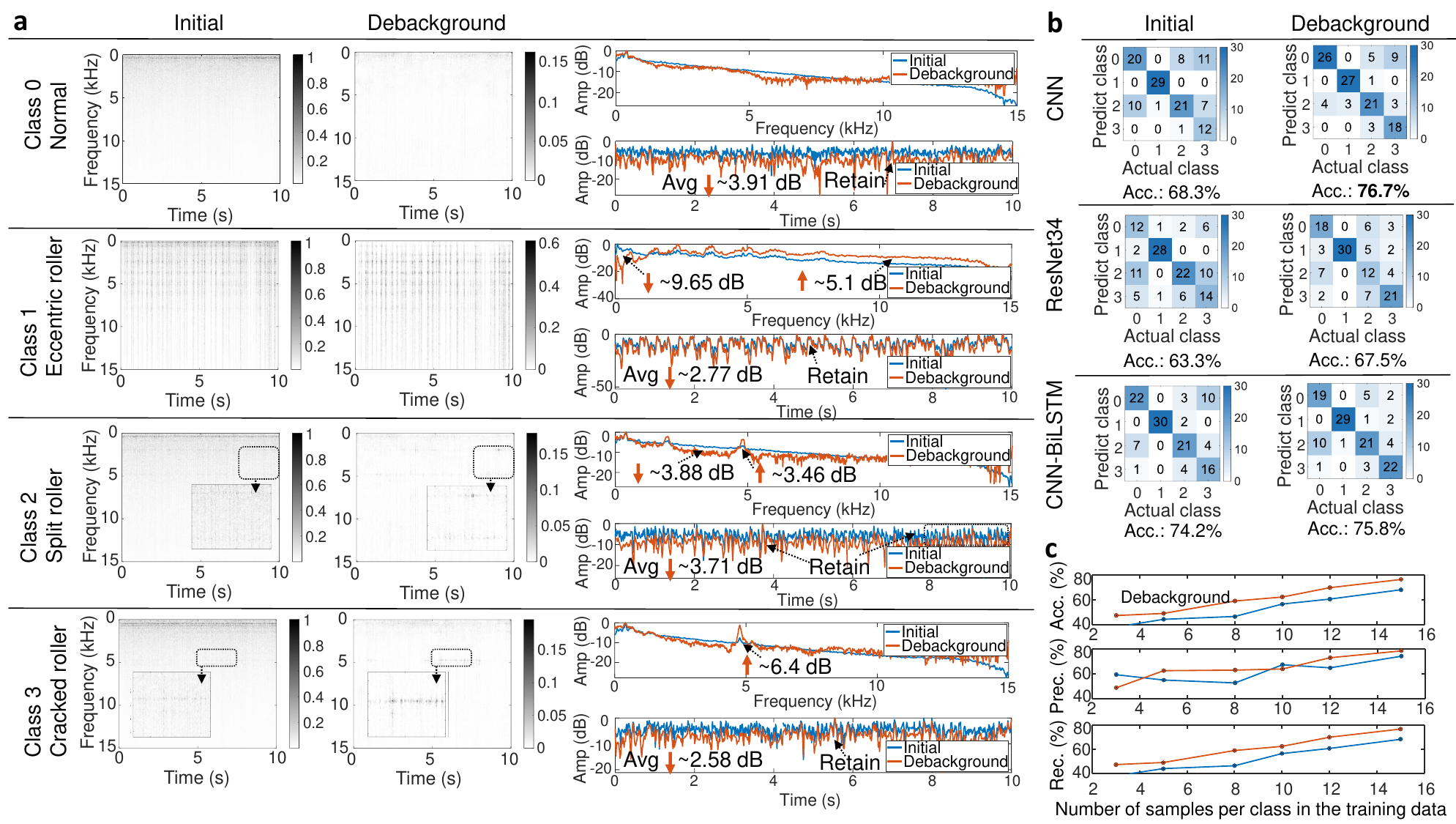}
\caption{\textbf{Debackground results from debackground net and fault diagnosis results in belt conveyor simulation test site.} \textbf{a} Comparison of initial and debackground data for different classes. Debackground net can eliminate background noise while retaining fault signals with sparse or broadband frequency feature. \textbf{b} Fault recognition results in different networks with initial and debackground data. Debackground data can achieve higher classification accuracy in various networks. \textbf{c} The classification performance of initial and debackground data in CNN under different training datasets with different sample numbers. Compared to the initial data, debackground data facilitates higher accuracy in classifier.}
\label{fig6}
\end{figure*}

The training and testing process of the proposed DAS debackground net are shown in Fig.4e and Fig.4f.
The first step is to obtain generated data from PIGN of fault events for training.
In conveyor belt fault monitoring, there is a discrepancy between theoretical models and field conditions, making effective physical modeling challenging.
Based on expert experience, roller faults in conveyor belts can be broadly categorized into sparse frequency and broadband frequency fault signals\cite{chen2020fault}.
By substituting the spatial dimension with the frequency dimension and employing expert experience functions as the objective feature functions (See in Methods), PIGN is capable of generating time-frequency data of roller fault (More results in Supplemental document), as shown in Fig.5.
With these generated data, debackground net learns to extract background data from the sum of generated data and real-world background data in the training process.
During testing, output background data of debackground net is firstly subtracted by input RWD, and then the result is passed through a ReLU activation function to acquire the event feature signal.
Debackground net adopts a residual learning structure, focusing on learning the background data rather than directly extracting event features (See in Methods).
The main reasons are the accessibility of a substantial amount of real-world background signals, and the low-frequency components within the background noise are more easily captured by the network, following the frequency principle in neural network training\cite{10.1007/978-3-030-36708-4_22}.
As shown in Fig.4d and Fig.4f, after the background noise is removed by debackground net, the subtle fault signal at frequency of 2 kHz can be distinctly observed.
The signal-to-noise ratio (SNR) is defined as the ratio of the fault signal intensity at 9 second to the value of the sum of the average and three times the root mean square from 5 to 8 seconds.
After removing the background noise, SNR has increased from 1.25 dB to 5.92 dB for this weak fault signal.

Experiment was implemented at belt conveyor simulation test site provided by Ningbo AllianStream Photonics Technology Co., Ltd, to acquire DAS data of normal roller, eccentric roller, split roller and cracked roller (See in Metohds).
The initial time-frequency data and the background removal results by debackground net for the four classes are shown in Fig.6a, with the temporal and frequency feature curves derived from the average energy calculations in the respective domains.
For the convenience of comparison, both the initial and debackground feature curves have been normalized individually.
After debackground, the maximum amplitude of the normal data decreased from 1 to 0.15, primarily retaining an abnormal feature at 7 second. 
The frequency feature curve of the normal data after debackground changed little, and the temporal feature curve showed an average decrease of $\sim$ 3.91 dB.
Fault data of eccentric roller with broadband frequency feature, as well as fault data of split roller and cracked roller with sparse frequency feature, are effectively stripped of background noise while retaining the fault signals after debackground net.
This is evident in the frequency feature curves where the intensities of fault frequencies are enhanced by $>$ 3dB, and in the time domain feature curve where it manages to retain the intermittent fault signals from similar background noise signals and achieves an average $>$ 2.5 dB reduction in background noise.
To demonstrate the benefits of debackground, both the initial and debackground data with label were used to train various classification networks under the same network structure and parameter settings.
The results with the same test dataset are shown in Fig.6b.
The training dataset includes 15 samples for each class, whereas the test data includes 30 samples for each class.
All networks were trained with $>$ 200 epochs to reach convergent state.
The results indicate that the debackground data demonstrate accuracy improvements of 8.4\%, 4.2\%, and 1.6\% in CNN, ResNet and CNN-BiLSTM respectively, when compared to the initial data.
Given the inherent randomness in the network training, multiple training rounds with different random parameter settings were performed.
After 10 rounds of training, the CNN achieved average accuracies of 74.42\% for debackground data and 70\% for initial data, with standard deviations of 1.71\% and 3.8\%, respectively.
This confirm the background noise removal capability of debackground net.
The debackground data shows performance improvement across different network structures, especially in simple network.
The performance of initial and debackground data are compared across small training datasets of varying sizes, as shown in Fig.6c.

\begin{figure*}[tb]
\centering
\includegraphics[width=17cm]{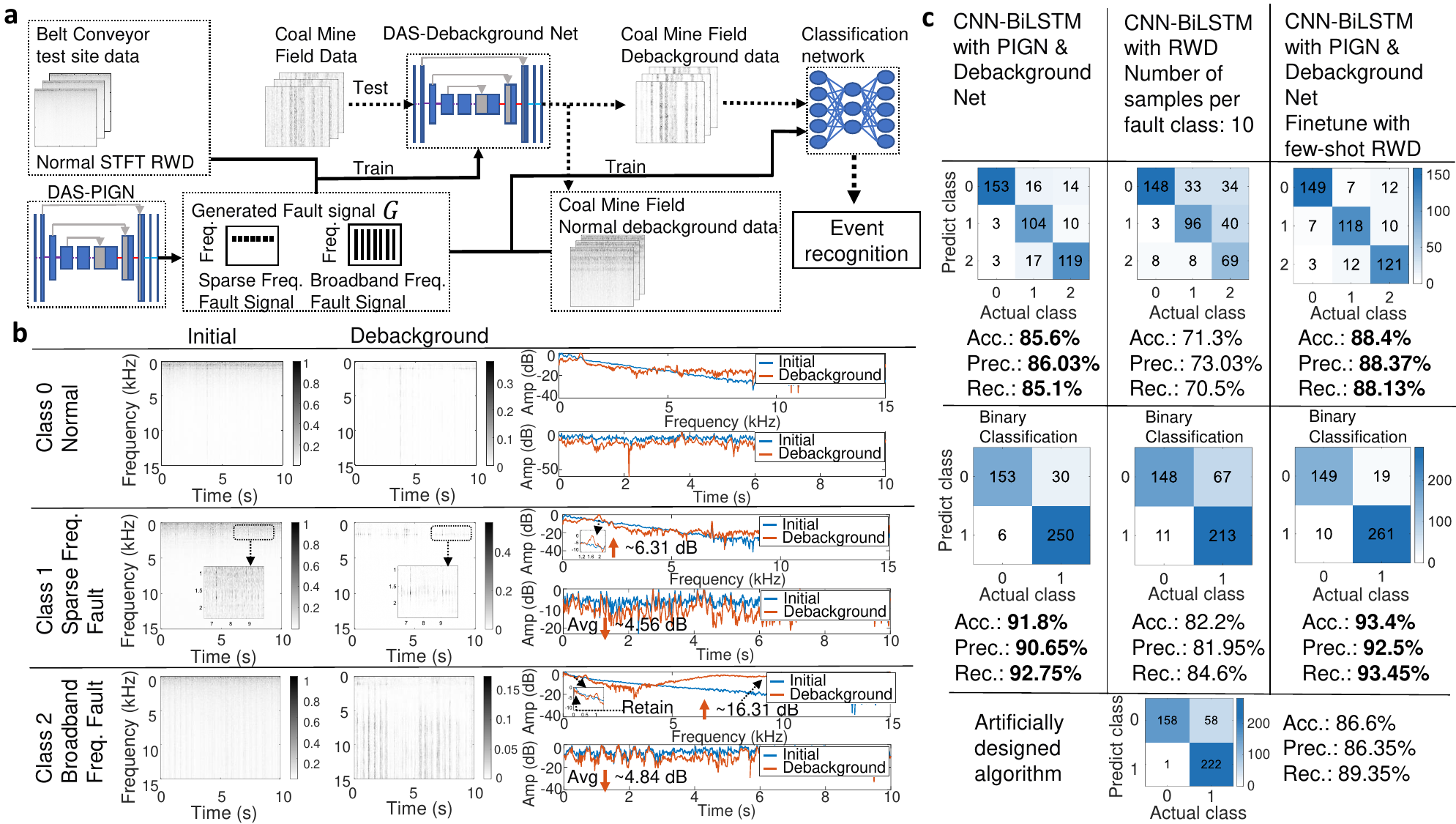}
\caption{\textbf{Deployment of physics-informed DAS neural network paradigm in coal mine field.} \textbf{a} Training and testing processes of PIGN, debackground net and classification network in coal mine field. Debackground net trained with background data from belt conveyor simulation test site and generated data from PIGN can be directly transferred to remove background noise in coal mine field data. The training of the classification network does not require fault data of simulation test or field. \textbf{b} Comparison of initial and debackground data in coal mine field of different classes. The debackground net, deployed directly from belt conveyor simulation test site to coal mine field, can still remove background noise and retain fault signals. \textbf{c} Fault recognition results of artificially designed algorithm and classification networks trained with generated debackground data or RWD. Class 0, normal; Class 1, sparse frequency fault; Class 2, broadband frequency fault. In binary classification task, class 0 and class 1 represent normal and fault, respectively. Physics-informed DAS neural network paradigm with PIGN and debackground net shows better classification performance than data-driven network and artifically designed algorithm.}
\label{fig7}
\end{figure*}

\subsection{Field test of belt conveyor fault detection}
In practical deployment of DAS, different sites of the same application often have distinct environments, which results in the poor generalization of traditional DAS data processing algorithm.
For a specific application, events to be detected across different sites follow the same physical model, and the corresponding DAS data should have similar features.
The main difference in DAS data for these similar events at different sites lies in the environmental background noise.
The proposed neural network paradigm that leverages event data generated from PIGN and easy-to-acquire background data for training facilitates the fast deployment in different fields.
Fig.7a illustrates the process of deploying a fault diagnosis network in a coal mine field of belt conveyor fault monitoring. 
The intrinsic mechanical vibration of belt conveyor constitutes the primary background noise.
Despite significant differences in the machine type, running speed, weight of load, and work environment of belt conveyor between the simulation test site and coal mine field, the background noise of the two sites still belongs to the vibration signal generated by mechanical rotating equipment.
Hence, the debackground net originally trained for the simulation test site can be directly applied to the coal mine field to remove background noise.
The debackground results of different types of events are shown in Fig.7b.
The debackground net remains effective in removing background noise from field data, reducing background noise by $>$ 4 dB in the time feature curves, while significantly enhancing the intensity of fault feature signals by over 6 dB in the frequency feature curves.
Even though the amplitude of background noise in the coal mine field is higher than the simulation test site, the denoising network still shows effective background noise removal capability.
The normal data of 450 samples collected in different time and space over a single day at the coal mine field are processed by the debackground network to obtain debackground real-world normal data.
Subsequently, these debackground real-world normal data were combined with fault signals generated by PIGN to constitute a training dataset for the classification network.
During a five-month field monitoring period, a total of 23 roller faults occurred on the belt conveyor. 
Approximately 10 samples were selected for each roller fault, combined with the normal data randomly selected in any time period during the 5 months, to form the test dataset.
The performance of CNN-BiLSTM trained with generated and debackground data on the test dataset is shown in Fig.7c.
For comparison, the same network structure was trained with labeled RWD as training data.
The real-world training dataset is an imbalanced dataset, in which 450 samples are in the normal clasee, and 10 samples are in each of the two fault classes.
Data augmentation of directly copying fault data and cost-sensitive learning methods are adopted to complete this unbalanced sample training task.
The proposed approach, without requiring any real-world fault data for training, achieves a three-class classification accuracy of 85.6\%, which is a 14.3\% improvement compared to the 71.3\% accuracy of the network trained with RWD.
In practical applications, it is usually only necessary to determine whether there is a fault.
Therefore, the binary classification results are presented, and additionally compared with the result of a complex manually designed fault diagnosis algorithm based on time-frequency analysis\cite{patent1}.
This algorithm was deliberately designed after a comprehensive review of data spanning five months and has undergone meticulous parameter adjustment.
Even with these manual efforts, the accuracy of artificially designed algorithm is 86.6\%, inferior to the accuracy of 91.8\% achieved by proposed paradigm.
In practical use, once a few real-world target event data is obtained, finetuning the network with these RWD can further enhance performance of fault diagnosis. 
After fine-tuning the pretrained network with labeled RWD consisting of 10 samples per fault class, the accuracy for three-class and two-class classification reached 88.4\% and 93.4\%, respectively.

\section{Dicussion}

The gap between features of PIGN generated data and RWD determines the performance of debackground net and the accuracy of event recognition within the proposed paradigm.
The key to achieve better performance is to fully consider the appropriate physical model and expert experience, and reasonably establish the functions of objective feature and physical constraints in PIGN.
In the cases presented in this paper, the physical models for shake and walk events are based on elementary mechanics and vibration theory, while the feature functions establishment for belt conveyor faults originates from the expertise of field workers and research summaries of fault signal characteristics with similar mechanical structure.
For shake and walk, more accurate models of forced vibration of ropes and pedestrian motion can be employed, respectively\cite{imanishi2009dynamic,xiang2010physics}. 
As for belt conveyor faults, signals obtained from traditional acoustic sensors can be regarded as expert experience, which can be analyzed to derive the fault feature functions in future.
For other DAS applications, it is feasible to establish objective feature functions by suitable theoretical models from relevant disciplines and empirical data from previous sensor deployments.
Then PIGN can be trained with these objective feature functions to generate corresponding target DAS events data.
For instance, mature simulation models of seismic waves and extensive data collected by conventional seismometers are available in the field of seismic wave monitoring\cite{virieux2009seismic,shearer1987surface}, which can be used for generation of seismic DAS data by PIGN.
The proposed paradigm can be applied to diverse DAS applications such as structural health monitoring, pipeline leak detection, natural hazard detectione and transportation monitoring, eliminating the reliance on real-world event data and realizing rapid deployment of application in different sites.

One of the biggest obstacles in DAS applications is that the DAS signal characteristics of the event in different applications or various sites are distinct, which makes DAS data processing algorithms usually can only resolve specific task at particular site.
The root cause is the diverse background noise signals and event signals in different work sites.
In the proposed network paradigm, by employing a debackground net and fully leveraging easily available background data, background noise signals can be effectively eliminated.
The physical models or expert knowledge usually allows for the representation of the basic shared features or significant distinctions of the same type of event across different sites.
However, it struggles to express the specific signal differences of the event at various sites.
Due to the shared characteristics of the event, the physics-informed DAS neural network can be directly deployed at different sites for the event identification.
To achieve a better performance at a specific site, it is still necessary to collect RWD for the neural network to learn the specific event signal feature in the site.
By adopting pre-training technology and combining it with self-supervised or semi-supervised approaches commonly found in data-driven neural network strategy, the network trained with generated data for general scenarios can further learn the subtle distinct signal features of a specific site from RWD.


The proposed network paradigm mainly focuses on the introduction of physical information and training methods.
In this article, three different classification network structures are tested to demonstrate that the proposed paradigm can improve classification performance in different classification network structures.
Due to the varying performance of different network structures on different types of data, appropriate classification network structures can be adopted based on the characteristics of data in practical DAS applications.
The proposed paradigm also includes a data generation network and a denoising network. 
Both networks in this article are based on the U-net structure. 
In theory, since the proposed paradigm does not alter the specific structure of the network, it can be executed on different data generation and denoising network structures.
Optimizing the structure of data generation, denoising and classification networks based on the characteristics of DAS data in different applications can bring better results.

In conclusion, the proposed physics-informed DAS neural network paradigm can generate massive DAS target events data according to physical information and has the ability to remove intense background noise.
PIGN is validated in two different DAS applications to generate different types of DAS events data, which reveals the potential for diverse DAS applications.
Classification networks are trained using generated DAS spatiotemporal data of shake and walk events and tested on a public dataset, which verifies the effectiveness of the generated data from PIGN.
In belt conveyor monitoring application with intense background noise, fault DAS time-frequency data are generated by PIGN for the training of DAS debackground net.
In the belt conveyor simulation test site, the classification accuracy of debackground data showed an 8.4\% improvement compared to the initial data with CNN network.
From the results of transferring the networks from the simulation testing site to the field, it shows rapid deployment capability across different sites of the same application of the proposed paradigm without the need for any prior acquisition of real-world target events data, and superior event recognition performance compared to data-driven neural networks and meticulously designed conventional algorithms.
After learning the physical knowledge in DAS event data, the network has improved its generalization and can adapt to different application sites, which is beneficial for covering current DAS application environments and exploring potential DAS applications.
Further combining small batch RWD for fine-tuning of the proposed network can continue to improve event recognition performance in specific application scenario.
The proposed paradigm, effectively overcoming the long-standing challenges of dependence on real-world target events data and the serious interference of environmental noise events in DAS applications, provides a physics-informed network solution for signal processing in fields of optical sensing, imaging and communication.

\section{Methods}

The details and supplements of the content in Methods, as well as more results, are in Supplementary document.

\textbf{Experiments}

The DAS system, ixDAS-4000, used for belt conveyor monitoring is provided by Ningbo AllianStream Photonics Technology Co., Ltd.
The working principle of ixDAS-4000 is based on time-gated digital optical frequency domain reflectometry (TGD-OFDR).
The maximum sensing range is 60 km. The minimum spatial resolution is 3.6 m. The self-noise level at 1 km is 10 $p\varepsilon /\sqrt {Hz}$ @20 Hz and 5 $p\varepsilon /\sqrt {Hz}$ @100 Hz.
When applying DAS to belt conveyor monitoring, it is necessary to lay the optical cable along the frame of the belt conveyor to ensure that the optical cable is in contact with the frame of the belt conveyor.
When a certain roller of the belt conveyor malfunctions, the accoustic waves generated by the malfunction will propagate through the frame and couple into the optical cable, which be detected by the DAS system.

The simulation test site for the belt conveyor is provided by Ningbo AllianStream Photonics Technology Co., Ltd.
The total length of the belt conveyor at the simulation test site is 13 m. The roller spacing is 1.2 m. Diameter of roller is 159 mm. The belt speed of the conveyor can be adjusted between 0 $\sim$ 5m/s. The belt conveyor is in an unloaded state during operation.
Three kinds of roller faults were manufactured manually, including eccentric, split and cracked roller.

The total length of belt conveyor at the coal mine field is 900 m. The roller spacing is $\sim$ 1.2 m. There are two types of belt conveyor roller diameters, 133 mm and 159 mm. When the belt conveyor is running, the coal mine are loaded. The data collection process lasted for about 5 months. Under the running state of the belt conveyor, the on-site environmental sound noise is strong with a intensity of $\sim$ 100 dBA measured by a sound meter (Model AWA 5653 Sound Level Meter).

\textbf{Physical model}

\medskip
\textbf{Physical model of DAS}

When the external accoustic wave is coupled to the optical fiber, stress is applied to the optical fiber to change its length, resulting in the phase change of light propagation on the optical fiber. 
The phase change of the Rayleigh backscattering light field on the fiber detected by DAS corresponds to the change of external strain or accoustic wave.
Therefore, it is necessary to reflect the force on the optical fiber or the change of the optical fiber length in the physical model of target events in DAS applications.

\medskip
\textbf{Shake event}

Shake can be regarded as the combination of forced vibration caused by external force and free vibration of optical fiber.
The position of a point on the optical fiber $x(t)$ can be expressed as:
\begin{equation}
x(t) = {A_1}{e^{ - \delta t}}\cos (\Omega t + \psi ) + B{F_0}\sin (\omega t - \phi )
\label{eq1.1}
\end{equation}
where ${A_1}$, $\delta $, $\Omega$ and $\psi$ are the initial amplitude, attenuation coefficient, vibration frequency and initial phase of free vibration; $B$, $\omega$ and $\phi$ are coefficient, vibration frequency and initial phase of forced vibration; $F_0$ is the initial amplitude of external force.
The change of fiber position can reflect the change of fiber length, which is equivalent to the change of DAS signal. A scaling factor $M(t)$ is used to approximate the process of fiber position change to DAS signal.
Therefore, the temporal feature function $T$ of shake event can be expressed as
\begin{equation}
T = M(t)x(t) + N(t)
\label{eq1.2}
\end{equation}
where $N$ is the noise, including random noise, DAS phase noise and fading noise.
The projection of shake data in space has no obvious features, so the spatial feature function $S$ can be expressed as
\begin{equation}
S = rand(s) + N(s)
\label{eq1.3}
\end{equation}
where $s$ is the spatial channel.
Because shake event is easy to make the signals on adjacent spatial channels of DAS have correlation, it is necessary to properly consider the spatial correlation constraint function when training PIGN for shake data generation.

\medskip
\textbf{Walk event}

Because the walking speed of pedestrians follows a sine function, the acceleration $a$ of pedestrians follows a cosine function, and can be expressed as:
\begin{equation}
a = \frac{{\pi {v_c}}}{c}\cos (\frac{{\pi t}}{c})
\label{eq1.5}
\end{equation}
where $v_c$ is amplitude of speed change, $c$ is angular velocity coefficient. 
For different walking modes, such as fast walking and slow walking, the values of $v_c$ and $c$ are different.
According to Newton's second law of motion, the force exerted by pedestrians on the ground can also be approximately considered as cosine variation and acts on the optical fiber laid to the ground.
Therefore, the temporal feature function $T$ of walk event can be expressed as:
\begin{equation}
T = {M_{fast}}(t){a_{fast}} + {M_{slow}}(t){a_{slow}} + N(t)
\label{eq1.5}
\end{equation}
where $M$ is the scaling factor. The function includes both fast and slow walking modes.
The spatial feature function modeling of the walk event is relatively complex, and here we simply approximate it with random numbers:
\begin{equation}
S = rand(s) + N(s)
\label{eq1.6}
\end{equation}

\medskip
\textbf{Belt conveyor fault event}

The faults in belt conveyors are mostly caused by rotating components such as rollers, so the temporal feature function of the fault signal can be approximately expressed as:
\begin{equation}
T = M(t)\sin (\frac{{2\pi }}{P}t + \varphi ) + N(t)
\label{eq1.7}
\end{equation}
where $M$ is the scaling factor, $P$ is rotational period of roller.
In the application of belt conveyor fault detection, it is difficult to achieve fault diagnosis through DAS spatiotemporal data, and it is necessary to perform separate time-frequency analysis on the signals at each spatial position. 
According to the reference on fault analysis of rotating machinery equipment and the experience of on-site workers\cite{chen2020fault}, the frequency feature of fault signals can be mainly divided into two classes: sparse frequency and broadband frequency.
To simplify the model, the single frequency of the fault signal in the frequency domain is approximated to the probability density curve of Gaussian distribution . 
Sparse frequency feature function can be expressed as:
\begin{equation}
F(f) = \sum\limits_{i = 1}^N {{A_i}\frac{1}{{{\sigma _i}\sqrt {2\pi } }}{e^{ - \frac{{{{(f - {\mu _i})}^2}}}{{2{\sigma _i}^2}}}}}  + N(f)
\label{eq1.8}
\end{equation}
where ${A_i}$, ${\mu _i}$ and ${\sigma _i}$ are the amplitude, center frequency and bandwidth of the $i$th fault frequency, respectively.
Broadband frequency feature function can be expressed as:
\begin{equation}
F(f) = rand(f) + N(f) + {A_0}
\label{eq1.8}
\end{equation}
where $A_0$ is the basic frequency amplitude value to ensure that the signal has signal at all frequencies for simulation of broadband signal.

\textbf{PIGN structure}

PIGN is based on U-net structure\cite{ronneberger2015u}. There are two main types of building blocks in PIGN: convolution block(3 $\times$ 3 convolution layer + Leaky ReLU + MaxPool) and up-convolution block (2 $\times$ 2 transpose convolution layer + Leaky ReLU).
The input and output data size of PIGN for generation of shake and walk data are 1000 $\times$ 12, while the size for belt conveyor fault generated data is 1025 $\times$ 584.
All the networks in this article are implemented on the Pytorch platform and a NVIDIA GeForce RTX 4090 is used for training.
PIGN has two training modes, namely untrained mode and trained mode.

In untrained mode, the parameters in the target feature functions are first taken within a reasonable range to obtain the corresponding target feature curves. PIGN only inputs a single random image to output an individual event data. During the training process, the square of the difference between the calculated feature curves of the network output data and the target feature curves will be used as the loss function to guide the network training. Therefore, in untrained mode, PIGN is more suitable for generating a small number of data or data with specific features. 
The whole training time with 10000 epochs for the generation of an individual event data is about 40 seconds in PIGN to generate shake and walk events, while takes about 95 seconds with 15000 epochs to generate an individual belt conveyor fault time-frequency data.

In trained mode, a large number of feature curves are firstly obtained based on the target feature function. During the PIGN training process, the input data consists of random noise images with the same number of target feature curves, corresponding labels are feature curves. In this mode, PIGN is considered to have learned the mapping relationship between random noise and target event data. The trained PIGN can generate distinct yet same type of event data when inputting different random noise images. The trained mode is more suitable for generating massive amounts of event data and general-purpose tasks.
The whole training time using training dataset with 1800 samples and 40000 epochs is about 7.2 hours in PIGN to learn the generation capability of shake and walk events, while takes about 10.6 hours using training dataset with 1024 samples and 2500 epochs to learn the generation capability of belt conveyor fault time-frequency data.
During the test, it takes 0.001 seconds to generate 16 samples of shake/walk data or 8 samples of belt conveyor fault data, respectively.

\textbf{Debackground net structure}

DAS debackground net is based on U-net structure. The network layers of debackground net are same as PIGN.
In the application of belt conveyor monitoring, 3960 samples of data with a size of 1025 $\times$ 584, which were composed of 100 normal background data from belt conveyor simulation test site combined with PIGN generated fault data, are used as input data for the debackground net training process, label data is the corresponding background data.
The training process of the network reached convergence after 150 epochs, taking a total time of $\sim$ 3 hours.
During the testing process, it only took about 0.003 seconds to simultaneously remove background noise from 8 samples of data.

The difference and comparison results between DAS debackground net and DAS feature extraction net are presented in the supplementary materials. In practical applications, the selection of DAS debackground net or DAS feature extraction net should be based on the frequency characteristics of the target event signal. Events with high frequency signals are more suitable for debackground net, while events with low-frequency signals require the selection of DAS feature extraction net.

\textbf{Classification networks structure}

Three types of networks are used as classification networks, namely CNN, ResNet, and CNN-BiLSTM.
These three types of neural networks are classic representative network structures in the development of neural networks.
Most network structures are based on these three types of networks.
CNN network is the basis of most current network structures.
The CNN network adopted in this article consists of two convolutional layers and one fully connected layer.
ResNet is the representative of deep network. The standard ResNet34 is adopted in the article.
CNN-BiLSTM is a classic work combining CNN network with other types of network. 
It can adaptively extract local and global features of data, and is suitable for spatio-temporal data processing.
CNN-BiLSTM adopted in this article consists of three convolutional layers, one BiLSTM layer and one fully connected layer.
These networks generally reach convergence after training 100 epochs.


\bigskip

\textbf{Acknowledgements}
This work is financially supported by National Natural Science Foundation of China (NSFC) under Grant No. 62405178, 62435004.

\textbf{Data, Materials, and Code Availability} 

The data that support the findings of this study are available from the corresponding author upon reasonable request.

\textbf{Author contributions} 
Y. Wan conceived the idea, performed the experiments, analyzed the results and prepared the manuscript. 
H. Wang analyzed the results.
X. Yu provided experimental equipment and assistance.
J. Chen, X. Fan and Z. He discussed the work and revised the paper.

\textbf{Conflict of interest} 

The authors declare no competing interests.


\bibliography{ref}   
\bibliographystyle{ieeetr}




\listoffigures

\end{spacing}
\end{document}